\documentclass[10pt, a4paper]{article}
\usepackage{lrec-coling2024}
\usepackage{multirow,diagbox,makecell,amssymb,bm}
\usepackage{amsmath,amsthm}
\usepackage{makecell}
\usepackage{graphicx}
\usepackage{pifont}
\usepackage{algorithm}
\usepackage{algorithmic}
\usepackage{amsfonts}       
\usepackage{booktabs}       
\usepackage{multirow}
\DeclareMathOperator*{\argmax}{argmax}

\newcommand*\Diff[1]

\title{
 VI-OOD: A Unified Representation Learning Framework for Textual Out-of-distribution Detection
 }

\name{Li-Ming Zhan$^{1}$,Bo Liu$^{1}$, Xiao-Ming Wu$^{1\dag}$\thanks{~$^\dag$ Corresponding author.}}
\address{Department of Computing, The Hong Kong Polytechnic University, Hong Kong S.A.R.$^1$\\ 
        \{lmzhan.zhan,bokelvin.liu\}@connect.polyu.edu.hk\\
         xiao-ming.wu@polyu.edu.hk\\}

\abstract{
Out-of-distribution (OOD) detection plays a crucial role in ensuring the safety and reliability of deep neural networks in various applications. While there has been a growing focus on OOD detection in visual data, the field of textual OOD detection has received less attention. Only a few attempts have been made to directly apply general OOD detection methods to natural language processing (NLP) tasks, without adequately considering the characteristics of textual data. 
In this paper, we delve into textual OOD detection with Transformers. We first identify a key problem prevalent in existing OOD detection methods: the biased representation learned through the maximization of the conditional likelihood $p(y\mid x)$ can potentially result in subpar performance. We then propose a novel variational inference framework for OOD detection (VI-OOD), which maximizes the likelihood of the joint distribution $p(x, y)$ instead of $p(y\mid x)$. VI-OOD is tailored for textual OOD detection by efficiently exploiting the representations of pre-trained Transformers. Through comprehensive experiments on various text classification tasks, VI-OOD demonstrates its effectiveness and wide applicability. Our code has been released at \url{https://github.com/liam0949/LLM-OOD}.
    \\ \newline  }
\begin{document}

\maketitleabstract

\section{Introduction}

Large-scale deep neural networks (DNNs) such as CNNs and Transformers, have brought about a revolutionary impact on numerous complex real-world machine learning applications.
Nevertheless, a notable drawback of DNNs remains their tendency to make \emph{overconfident} decisions, rendering them less reliable for safety-critical applications like medical diagnosis~\cite{ulmer2020trust} and self-driving cars~\cite{DBLP:conf/icml/FilosTMRLG20}.
It has been noted that DNNs often assign elevated confidence scores to unfamiliar inputs, leading to potential erroneous predictions when confronted with anomalous out-of-distribution (OOD) data~\cite{DBLP:conf/cvpr/NguyenYC15}.
To address this issue, there has been active research and investigation into OOD detection in recent years~\cite{DBLP:conf/icml/HendrycksBMZKMS22,yang2022openood}.

\textbf{Challenge of OOD detection.} OOD detection aims at solving a ${K}$-class in-distribution (ID) classification task and a binary ID \emph{vs.} OOD discrimination task simultaneously. A commonly assumed practical setting is OOD examples are unavailable during training, which presents the major challenge for OOD detection. The mainstream methods for OOD detection commonly follow a post-hoc scheme~\cite{DBLP:conf/iclr/HendrycksG17}, 
which first discriminatively trains an ID $K$-class classifier by maximizing the conditional likelihood of $p(y \mid x)$ and then derives some statistics from the trained model to predictive OOD confidence scores. However, since the binary ID \emph{vs.} OOD discrimination task is not considered in the training process, the learned representations by $K$-class training may be biased to the ID classes. While some attempts have been made to address this challenge by incorporating surrogate OOD datasets during the training phase, such as those described in the works by \citet{DBLP:conf/iclr/HendrycksMD19} and \citet{ DBLP:conf/iclr/LeeLLS18}, further endeavors are required to identify appropriate OOD datasets that demonstrate significant distributional shifts compared to the ID data.


\textbf{Research on textual OOD detection.} The majority of recent research efforts have concentrated on detecting OOD data in visual domains, with only a limited number of studies~\cite{hendrycks2020pretrained, DBLP:conf/aaai/PodolskiyLBAP21,DBLP:conf/emnlp/Zhou0C21} focusing on textual OOD detection. As far as our knowledge extends, current textual OOD detection methods typically utilize general OOD detection algorithms on representations generated by Transformers~\cite{vaswani2017attention}. However, these methods often fail to adequately account for the unique characteristics and nuances of textual data. Moreover, although the hierarchical contextual representations of pre-trained Transformers have demonstrated remarkable effectiveness in numerous NLP tasks~\cite{DBLP:conf/cncl/SunQXH19,DBLP:journals/corr/abs-1910-07973, DBLP:conf/emnlp/MohebbiMP21, DBLP:conf/naacl/DevlinCLT19, DBLP:journals/corr/abs-1907-11692}, their potential for textual OOD detection has not been fully harnessed.



\textbf{Our proposal.} To tackle the aforementioned issues, we propose a variational inference framework based on Transformers for textual OOD detection.
Rather than solely focusing on maximizing the conditional distribution $p(y|x)$ of ID data, our approach involves optimizing the joint distribution $p(x,y)$, which is to maximize $p(y|x)$ and $p(x)$ simultaneously. The core idea revolves around modeling the distribution of the provided ID data, which allows us to harness valuable information that might not be directly relevant for ID classification but proves significant for outlier detection.
To make the joint distribution $p(x,y)$ tractable, we resort to optimizing the evidence lower bound of $p(x,y)$ derived via amortized variational inference (AVI)~\cite{DBLP:journals/corr/KingmaW13}. 
Moreover, considering the unique characteristics of textual data, we modify the approximated posterior distribution in the framework of AVI, making the posterior conditioned on a dynamic combination of intermediate layer-wise hidden states of the Transformer.
The Transformer backbone functions as a shared encoder for both the ID classification head and the decoder (generator) in the AVI framework (Fig.~\ref{fig:model}).

The contributions of this work include:
\begin{itemize}


 \item Our proposed variational inference framework for OOD detection (VI-OOD) offers a novel and principled approach, providing a fresh perspective that is orthogonal to previous OOD detection methods~\cite{hendrycks2020pretrained, DBLP:conf/aaai/PodolskiyLBAP21,DBLP:conf/emnlp/Zhou0C21}.

 \item Our instantiation of VI-OOD harnesses the rich contextual representations of pre-trained Transformers to learn more effective latent representations for text inputs. The improved representations can be readily used by various existing post-hoc OOD detection algorithms, consistently enhancing their performance in textual OOD detection.

 \item Our proposed method is evaluated using mainstream encoder-based and decoder-based Transformer architectures and comprehensive OOD text classification scenarios. It can offer advantages to widely utilized OOD detection algorithms, particularly for distance-based OOD detectors, such as the Mahalanobis Distance method~\citep{DBLP:conf/nips/LeeLLS18}.
\end{itemize}






\section{Pilot Study}\label{sec:motivation}


 \begin{figure*}[ht]
    \centering
    \scalebox{1.}{
    \includegraphics[width=1.0\linewidth]{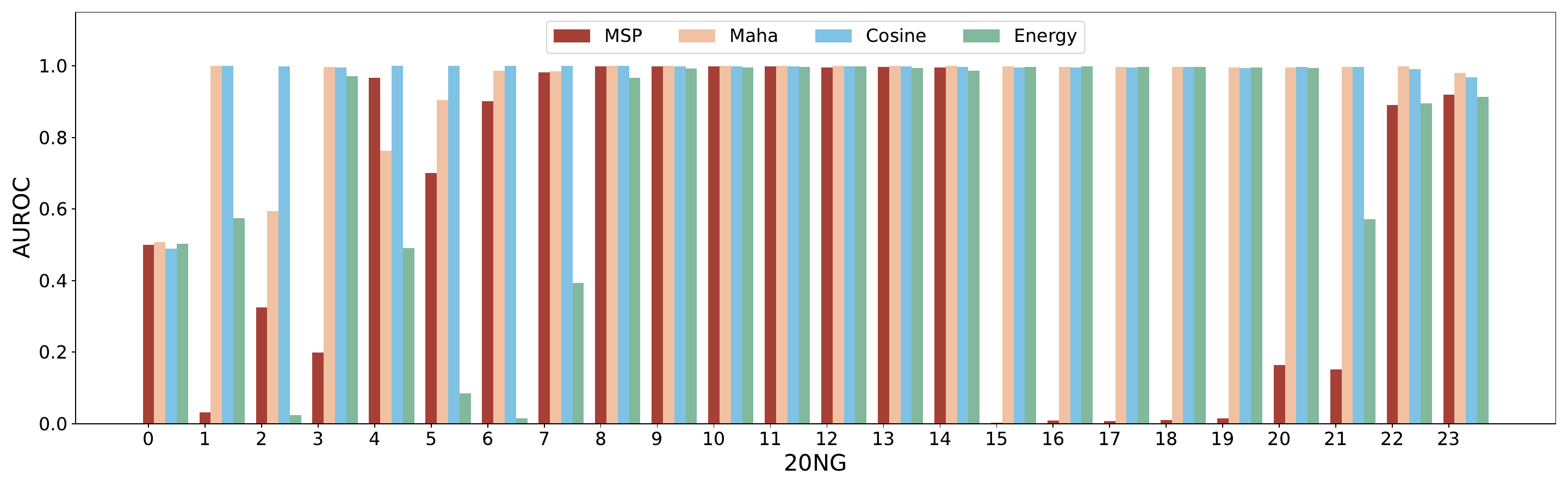}
    }
    \caption{
    Investigation of OOD performance of Transformer's intermediate Hidden States: AUROC Results for 24 Layers of RoBERTa\textsubscript{LARGE}. The figure illustrates the OOD performance evaluation across multiple layers of RoBERTa\textsubscript{LARGE}. Higher values indicate better performance. The model undergoes fine-tuning on SST-2 and is assessed for OOD performance using the 20NG dataset. The four commonly used OOD scoring functions, namely MSP (red), Maha (light yellow), Cosine (blue), and Energy (green), are represented in the figure.
    }
    \label{fig:layers}
\end{figure*}
\subsection{Problem Statement and Motivation}
\textbf{Out-of-distribution (OOD) detection} aims to accurately separate all class-dependent in-distribution (ID) examples as well as out-of-distribution (or anomalous) examples.
Given the input space $\mathcal{X} \times \mathcal{Y}$ and an ID class label set $\mathcal{Y}_{\text{ID}}= \{y_j\}_{j=1}^{K} \subset \mathcal{Y}$, an ID training set $\mathcal{D}_{\text{ID}}=\{(x_i, y_i)\}_{i=1}^{N}$ is sampled from the distribution $p(x, y)$ of ID data where $ y_i\in \mathcal{Y}_{\text{ID}}$. 
With $\mathcal{D}_{\text{ID}}$, an ID classifier $f_\text{ID}: \mathcal{X} \rightarrow \mathcal{Y}_\text{ID}$ is trained.
During test time, since there may be a distribution shift between the training and test data in practical application scenarios~\cite{DBLP:journals/corr/SzegedyZSBEGF13,DBLP:conf/aistats/MorningstarHGLA21},
the ID classifier $f_\text{ID}$ may encounter OOD samples ($y_i \notin \mathcal{Y_\text{ID}}$). Hence,
an OOD confidence scoring function $f_\text{OOD}: \mathcal{X} \rightarrow \mathbb{R}$ is needed to perform ID \emph{vs.} OOD binary classification. 
In this regard, OOD detection aims to solve both the $K$-class ID classification task and the binary outlier detection task. The ID classifier $f_\text{ID}$ is commonly trained with a discriminative loss by maximizing the conditional log-likelihood of the training set:
\begin{flalign}
\begin{aligned}\label{eq:cross_loss}
 \hat{\mathbf{\theta}} = \argmax_{\mathbf{\theta}}\frac{1}{N}\sum_{(x_i, y_i) \in \mathcal{D}_{\text{ID}}} \log p(y_i\mid x_i; f_\text{ID},\mathbf{\theta}),
\end{aligned}
\end{flalign}
where $\mathbf{\theta}$ stands for all trainable parameters of $f_\text{ID}$.

\textbf{The fundamental challenge of} OOD detection is that at the training stage, real OOD examples are unavailable and thus cannot be effectively represented to provide necessary learning signals for the binary ID \emph{vs.} OOD task. To address this issue, a few attempts have been made to introduce surrogate OOD datasets during training by using some datasets irrelevant to the ID data ~\citep{DBLP:conf/iclr/HendrycksMD19,DBLP:conf/iclr/LeeLLS18}. 
However, it is difficult to select suitable ``OOD'' datasets to represent the huge space of real OOD data.

\textbf{Post-hoc methods.} The majority of existing OOD detection methods~\citep{DBLP:conf/iclr/HendrycksG17, DBLP:conf/iclr/HendrycksMD19, DBLP:conf/nips/LeeLLS18, DBLP:conf/nips/LiuWOL20, DBLP:conf/icml/HendrycksBMZKMS22, DBLP:conf/nips/SunGL21,DBLP:conf/icml/SunM0L22} follow a post-hoc paradigm and address the binary ID \emph{vs.} OOD task in the inference stage.
These methods propose different OOD confidence scoring functions with the trained ID classifier $f_\text{ID}$. 
Specifically, 
the parameters of the trained $f_\text{ID}$ are frozen, and some statistics of specific layers of $f_\text{ID}$ (usually the penultimate layer or the softmax layer) are often used as OOD confidence scores. 

\textbf{Motivation of this work.}
While post-hoc methods have shown promise, it is pointed out that the performance of $f_\text{ID}$ on ID data is not a good indicator of its performance on OOD data~\cite{hendrycks2020pretrained, DBLP:conf/iclr/LeeLLS18}. Specifically, the discriminative training of $f_\text{ID}$ is often conducted with $p(y|z)$, where $z$ is the latent representation obtained by passing an input $x$ to a DNN encoder.
Maximizing the conditional log-likelihood $\log p(y|z)$ is essentially maximizing the mutual information between the latent variable $Z$ and the label variable $Y$, i.e., $\mathcal{I}(Z, Y)$~\cite{boudiaf2020unifying}. Naturally, the learned representation $Z$ will be biased towards the ID classification task. Indeed, \citet{DBLP:journals/corr/abs-2003-00402} have demonstrated that in the Mahalanobis-distance-based OOD detection method, the principal components of ID data that are deemed least important for the ID classification task actually contain valuable information for the binary ID \emph{vs.} OOD task. This information may be overlooked or discarded when training the ID classification function $f_\text{ID}$ using the conditional likelihood $p(y|x)$. Furthermore, a recent study by \citet{uppaal2023fine} highlights that relying solely on supervised training with ID data can lead to a degradation in the performance of OOD detection as the training progresses.


To address this issue, we propose to learn better latent representation $Z$ for post-hoc methods by considering the distribution of ID data, i.e., maximizing the likelihoods $p(y|x)$ and $p(x)$ simultaneously\footnote{Note that $p(y|x) = \int_{z}p(y|z, x) p(z|x) \, dz$  and $p(x)=\int_{z}p(x|z)p(z) \, dz$.}, which is equivalent to modeling $p(x, y)$ -- the joint distribution of ID data. To this end, we design a novel principled variational framework that will be elaborated in the next section. 

\subsection{A Closer Look at Textual OOD Detection with Transformers}
In Figure~\ref{fig:layers}, we study the impact of the intermediate hidden states of RoBERTa\textsubscript{LARGE} on textual OOD detection. Following \cite{hendrycks2020pretrained}, we take the model trained on SST-2 as a case study. The model is trained solely with the discriminative loss. 
We conduct OOD detection by utilizing each hidden state of the trained model's $24$ layers as representations of the input text data. Subsequently, we summarize the AUROC results obtained from four commonly used OOD detection algorithms. 
As the layer number increases from $0$ to $23$, the hidden layer is closer to the head of the model, i.e., layer $23$ outputs the last hidden state. 

\begin{figure*}[ht]
    \centering
\includegraphics[width=1.0\linewidth]{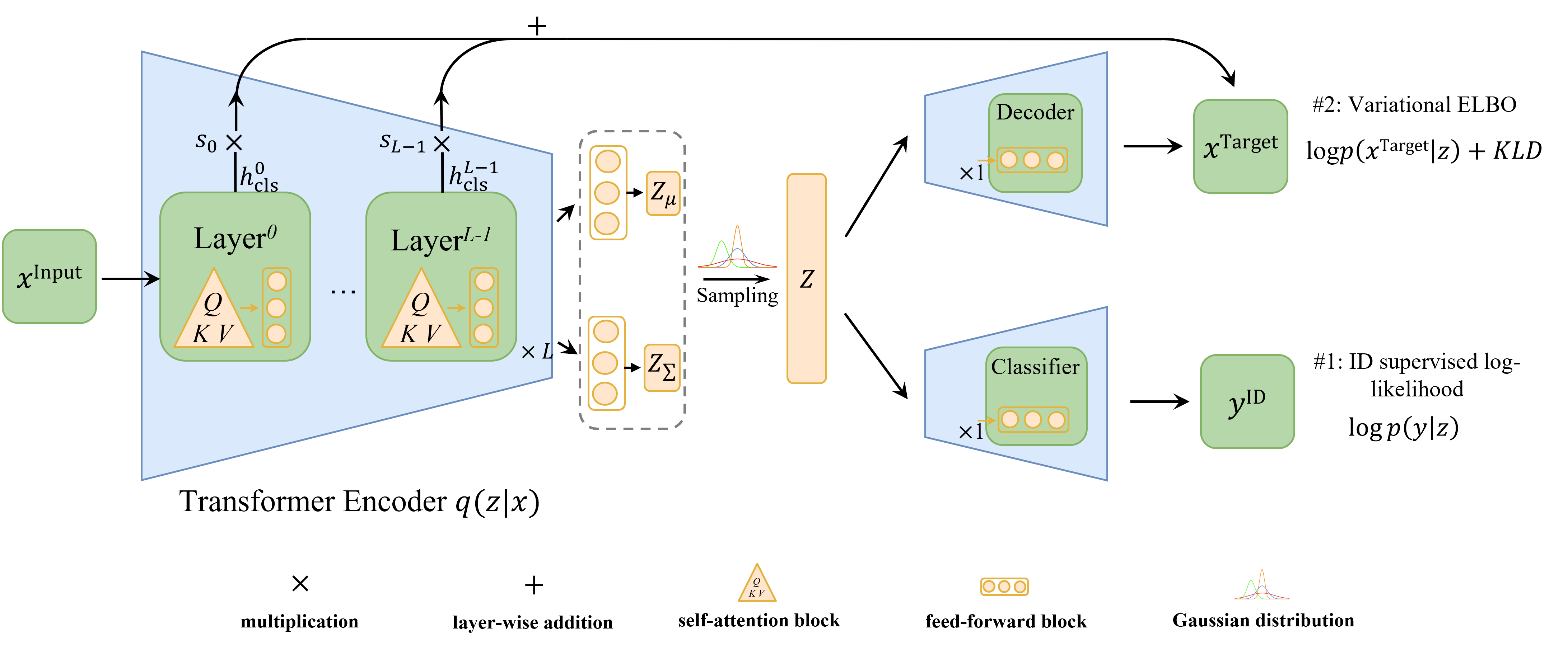}
\caption{The architecture of our proposed framework. Our method employs an encoder-based transformer model as the backbone textual encoder. Hidden states of the [CLS] token are chosen to be textual representations. $z$ is a latent variable conditioned on the textual representations. The in-distribution (ID) classification head $p(y|z)$ and decoder $p(x^{\text{target}}|z)$ both take $z$ as the input. $\mathbf{s}$ is the hidden states combination factor and the merge representation $x^{\text{target}}$ works as the target of the decoder.
}

\label{fig:model}
\end{figure*}
\paragraph{Intermediate hidden states could help OOD detection.} The results presented in Figure~\ref{fig:layers} clearly indicate that intermediate hidden states consistently outperform the final hidden states in terms of OOD performance, as observed across all four OOD detection methods. The best performance consistently occurs in the middle layers, particularly in the range of layers 9 to 13. This consistent performance is observed for all four OOD detection methods.
On the other hand, as pointed out by \citet{DBLP:conf/cncl/SunQXH19}, intermediate hidden states of Transformers exhibit inferior performance compared to the final hidden state in ID classification tasks. Based on these observations, we make a key assumption: \textbf{intermediate hidden states contain redundant information for ID classification but crucial information for OOD detection.}


Furthermore, it is possible to address the disparities among various OOD detection methods. As depicted in Figure~\ref{fig:layers}, the performance of intermediate layers (layers 9 to 14) is consistently comparable across the four OOD detection methods. For instance, the Maximum Softmax Probability (MSP) method demonstrates excellent results around layer 13, but its performance significantly deteriorates at the last layer, layer 23. These findings suggest that effectively harnessing the potential of hidden states in Transformers can alleviate the challenges associated with OOD detection.



\section{Proposed Method}

\subsection{VI-OOD: A Variational Inference Framework for Out-of-distribution Detection}
Our goal is to directly maximize the likelihood of the joint distribution $p(x, y)$ rather than $p(y|x)$. 
We assume that a latent variable $Z$ is a stochastic encoding of the input sequence $X$. The log likelihood of $p(x, y)$ can then be calculated by:
\begin{align}\label{eq:factor}
\log p(x, y)
& = \log \int_{z} p(x, y, z) \, dz  \nonumber\\
&= \log \int_{z} p(y|z,x)p(x|z)p(z) \, dz \nonumber\\
& = \log \int_{z} p(y|z)p(x|z)p(z) \, dz,
\end{align}
where in the last equality we assume the Markov chain $X \leftrightarrow Z \leftrightarrow Y$, i.e., $p(y|z,x) = p(y|z)$. 
Since it is intractable to compute the integral in Eq.~(\ref{eq:factor}), we employ amortized variational inference~\cite{DBLP:journals/corr/KingmaW13} to derive the lower bound of $\log p(x, y)$ as follows.
\begin{align}
\log p(x, y)
&=  \log \int_{z} p(y|z)p(x|z)p(z) \, dz \nonumber\\
&= \log \int_{z} p(y|z) p(x|z) p(z) \frac{q(z|x)}{q(z|x)} \, dz \label{eq:avae} \\
&= \log \mathbb{E}_{z\sim q(z|x)} \left[\frac{p(y|z) p(x|z) p(z)} {q(z|x)}\right] \\
&\geq \mathbb{E}_{z\sim q(z|x)}\left[ \log \frac{p(y|z) p(x|z) p(z)} {q(z|x)} \right], \label{eq:ajesen}
\end{align}
where $q(z|x)$ in Eq.~(\ref{eq:avae}) is the amortized variational approximator of the true posterior $p(z|x)$, and Jensen’s inequality is applied in Eq.~(\ref{eq:ajesen}).
The last quantity in Eq.~(\ref{eq:ajesen}) is the evidence lower bound of $\log p(x, y)$, which can be rewritten as:
\begin{align}
\label{eq:final_loss}
\mathcal{L}_\text{ELBO}
= &\underbrace{\mathbb{E}_{z}\left[\log p(y|z) \right]}_{\text{Target \#1: ID supervised training}} +\nonumber \\
&\underbrace{\mathbb{E}_{z}\left[\log p(x|z) \right]- D_\text{KL}(q(z|x)||p(z))}_{\text{Target \#2: Unsupervised variational training}},
\end{align}
where the first term is the ID supervised training objective, and the second and third terms correspond to the unsupervised learning objective for an amortized variational Bayesian autoencoder.


\subsection{Transformer-based Textual OOD Detection with VI-OOD}

Our proposed VI-OOD framework is a general probabilistic approach for learning data representations, which can be applied to various types of data, including image, textual, audio, and video. However, in this work, we focus on textual data. In the following, we outline the instantiation of VI-OOD for textual OOD detection, which involves designing the encoder (posterior approximator) $q(z|x)$, the decoder (reconstructor) $p(x|z)$, and the discriminator $p(y|z)$, as depicted in Figure~\ref{fig:model}.





\paragraph{Encoder for learning textual representations.} 
Encoder-based Transformers have become a prevailing standard in learning contextual representations of text due to their excellent performance in numerous NLP tasks. Hence, the transformer architecture is a natural choice for the encoder $q(z|x)$. In this paper, we utilize models from the BERT family~\cite{DBLP:conf/naacl/DevlinCLT19}. 
Given an input $x$, which is a sequence of tokens with a length of $N$, denoted as $[x_{0},\cdots ,x_{N-1}]$, BERT adds a special token [CLS] at the start of the input sequence,
i.e., $[\text{CLS}, x_{0},\cdots ,x_{n-1}]$. The inclusion of the [CLS] token is intended for classification tasks. Unless otherwise specified, we use the hidden states of the [CLS] token as the textual representations. The input sequence $x$ is passed through each layer of BERT, resulting in a series of intermediate hidden states at the [CLS] position, denoted as $h_\text{CLS}=[h_\text{CLS}^{0}, \cdots, h_\text{CLS}^{L-1}]$, where $L$ is the total number of layers. As shown in Figure~\ref{fig:model}, we instantiate the encoder $q(z|x)$ and the prior $p(z)$ as diagonal Gaussian distributions, i.e., $\mathcal{N}(z |\mu, \Sigma)$ and $\mathcal{N}(0, I)$ respectively, where $\mu$ and $\Sigma$ are obtained by mapping the last hidden state $h_\text{CLS}^{L-1}$ with a single-layer MLP respectively. 


\paragraph{Decoder for reconstructing the textual representations.} 
In the case of image data, selecting the original input image as the decoder target for reconstruction is straightforward since it contains the most informative content. However, when working with textual data, the input token sequence only represents embeddings from a predefined dictionary, while the intermediate hidden states of the Transformer capture valuable contextual semantics. As a result, determining the appropriate reconstruction target for $p(x|z)$ in textual data poses a challenging task. To leverage the potential of the intermediate hidden states, our approach aims to condition the reconstruction target on the hidden states. Based on our preliminary experiments, we observed that different hidden layers have varying effects on different ID datasets. Consequently, it is difficult to predefine a fixed combination pattern for integrating the intermediate hidden states.
Therefore, we introduce a learnable weight vector $\mathbf{s} = [s_0, \cdots, s_{L-1}]\in \mathbb{R}^{L}$ to dynamically integrate the intermediate hidden states of the Transformer. 
Then, we derive the reconstruction target: 
\begin{align}
    x^{\text{target}}= (h_\text{CLS}^{0} \cdot s_0) + (h_\text{CLS}^{1} \cdot s_1) + \cdots (h_\text{CLS}^{L-1} \cdot s_{L-1}),\nonumber
\end{align}
where $\cdot$ denotes multiplication. In this way, $x^{\text{target}}$ contains rich contextualized semantic information. Referring to Figure~\ref{fig:model}, we realize the reconstructor (decoder) $p(x|z)$ as a single feed-forward block, taking a sample $z$ from $\mathcal{N}(z |\mu, \Sigma)$ as input and outputting a reconstructed version of $x^\text{target}$ to maximize $p(x^{\text{target}}|z)$.
The ID classifier $f_{\text{ID}}$ is a single-layer MLP that takes the latent representation $z$ as input.





\paragraph{Discriminator for ID classification and binary OOD detection.}
At the inference stage, we only need the trained posterior approximator (encoder) $q(z|x)$ and the ID classifier $f_\text{ID}$. Note that both the ID classification task and the binary outlier detection task are performed w.r.t. the latent variable $z$. For each $x$, we only sample one $z$ during training and inference respectively.

\section{Related Work}

\paragraph{OOD detection based on density estimation.} 

Besides the problem setting discussed in Section~(\ref{sec:motivation}), 
another line of works tries to address the OOD detection problem by solving a more general problem -- density estimation.
Unlike the setting of our work, the focus of these works is solely on the binary classification task of distinguishing between in-distribution (ID) and OOD samples, disregarding the ID classification task.
Their learning target is the density function of the training set -- $p_{ID}(x)$ -- such that OOD examples are assumed to yield lower probabilities than the ID ones. However, in high dimensional spaces, this assumption is not held in practice and many previous works~\cite{choi2018waic} have found that OOD examples may be assigned higher likelihoods than ID examples. Recent works~\cite{ren2019likelihood,DBLP:conf/iclr/NalisnickMTGL19, DBLP:conf/aistats/MorningstarHGLA21} are still trying to correct this pathology. 


In particular, numerous prior studies have leveraged the density estimation capabilities of variational autoencoders (VAEs) for OOD detection. For instance, \citet{floto2023the} enhance VAEs for OOD detection by substituting the standard Gaussian prior with a more versatile tilted Gaussian distribution. Likelihood Regret~\cite{xiao2020likelihood} and Likelihood ratios~\cite{ren2019likelihood} adopt a similar perspective of training two distinct models--one capturing the semantic content of the data, and the other capturing background information. Their major difference is the training data of the background model and semantic model.

\paragraph{OOD detection in NLP.}
OOD detection in the NLP domain has recently attracted increased attention~\cite{liu2023good}. OOD intent detection~\cite{zhang2021deep, zhan-etal-2021-scope} investigates the OOD detection problem for anomalous utterances in dialogue systems. \citet{podolskiy2021revisiting} empirically find out that Mahalanobis Distance is the best performing OOD scoring function for OOD intent detection.
A few attempts has been made to study the general textual OOD detection problem. \citet{hendrycks2020pretrained} point out that pre-trained Transformers are more robust for OOD detection than previous model architectures~\cite{hochreiter1997long}. \citet{zhou2021contrastive} and \citet{cho2022enhancing} employ a contrastive regularizer to learn better representations for textual OOD detection. \citet{uppaal2023fine} conduct an evaluation on RoBERTa and point out that ID fine-tuning may pose a detrimental effect on textual OOD detection.

\section{Conclusion}
This paper concentrates on exploring Out-of-Distribution (OOD) detection within Natural Language Processing (NLP) classification tasks using Transformer-based large language models (LLMs). Building on our detailed analysis of hidden states in Transformers, we introduce a Variational Bayesian framework named VI-OOD. This framework optimizes the joint distribution $p(x, y)$ during the training phase. Our methodology is reinforced by both experimental evidence and theoretical analysis, underscoring its validity. We have rigorously tested our approach with mainstream Transformer architectures, encompassing both encoder-based and decoder-based models. Comprehensive experiments on diverse textual classification tasks affirm the efficacy and superiority of our OOD detection framework.

This research is dedicated to enhancing AI safety and the robustness of models. As such, our findings are poised to benefit various AI applications without presenting a direct risk of misuse. Furthermore, our proposed methodology relies exclusively on open-source benchmarks for training data, avoiding the introduction of additional datasets for training the OOD detector. As such, our approach sidesteps potential ethical concerns associated with data collection. Additionally, by building upon open-source LLMs, our method avoids substantial increases in resource consumption, aligning with principles of sustainable and responsible AI development.

\section*{Acknowledgments}
We thank the anonymous reviewers for their helpful feedback. This research was partially supported by the grant of HK ITF ITS/359/21FP.

\section{References}\label{sec:reference}

\bibliographystyle{lrec-coling2024-natbib}
\bibliography{lrec-coling2024}

\begin{thebibliography}{39}
\expandafter\ifx\csname natexlab\endcsname\relax\def\natexlab#1{#1}\fi

\bibitem[{Boudiaf et~al.(2020)Boudiaf, Rony, Ziko, Granger, Pedersoli, Piantanida, and Ayed}]{boudiaf2020unifying}
Malik Boudiaf, J{\'e}r{\^o}me Rony, Imtiaz~Masud Ziko, Eric Granger, Marco Pedersoli, Pablo Piantanida, and Ismail~Ben Ayed. 2020.
\newblock A unifying mutual information view of metric learning: cross-entropy vs. pairwise losses.
\newblock In \emph{Computer Vision--ECCV 2020: 16th European Conference, Glasgow, UK, August 23--28, 2020, Proceedings, Part VI}, pages 548--564. Springer.

\bibitem[{Cho et~al.(2022)Cho, Park, Kang, Yoo, Kim, and Lee}]{cho2022enhancing}
Hyunsoo Cho, Choonghyun Park, Jaewook Kang, Kang~Min Yoo, Taeuk Kim, and Sang-goo Lee. 2022.
\newblock Enhancing out-of-distribution detection in natural language understanding via implicit layer ensemble.
\newblock \emph{arXiv preprint arXiv:2210.11034}.

\bibitem[{Choi et~al.(2018)Choi, Jang, and Alemi}]{choi2018waic}
Hyunsun Choi, Eric Jang, and Alexander~A Alemi. 2018.
\newblock Waic, but why? generative ensembles for robust anomaly detection.
\newblock \emph{arXiv preprint arXiv:1810.01392}.

\bibitem[{Devlin et~al.(2019)Devlin, Chang, Lee, and Toutanova}]{DBLP:conf/naacl/DevlinCLT19}
Jacob Devlin, Ming{-}Wei Chang, Kenton Lee, and Kristina Toutanova. 2019.
\newblock \href {https://doi.org/10.18653/v1/n19-1423} {{BERT:} pre-training of deep bidirectional transformers for language understanding}.
\newblock In \emph{Proceedings of the 2019 Conference of the North American Chapter of the Association for Computational Linguistics: Human Language Technologies, {NAACL-HLT} 2019, Minneapolis, MN, USA, June 2-7, 2019, Volume 1 (Long and Short Papers)}, pages 4171--4186. Association for Computational Linguistics.

\bibitem[{Filos et~al.(2020)Filos, Tigas, McAllister, Rhinehart, Levine, and Gal}]{DBLP:conf/icml/FilosTMRLG20}
Angelos Filos, Panagiotis Tigas, Rowan McAllister, Nicholas Rhinehart, Sergey Levine, and Yarin Gal. 2020.
\newblock \href {http://proceedings.mlr.press/v119/filos20a.html} {Can autonomous vehicles identify, recover from, and adapt to distribution shifts?}
\newblock In \emph{Proceedings of the 37th International Conference on Machine Learning, {ICML} 2020, 13-18 July 2020, Virtual Event}, volume 119 of \emph{Proceedings of Machine Learning Research}, pages 3145--3153. {PMLR}.

\bibitem[{Floto et~al.(2023)Floto, Kremer, and Nica}]{floto2023the}
Griffin Floto, Stefan Kremer, and Mihai Nica. 2023.
\newblock \href {https://openreview.net/forum?id=YlGsTZODyjz} {The tilted variational autoencoder: Improving out-of-distribution detection}.
\newblock In \emph{The Eleventh International Conference on Learning Representations}.

\bibitem[{Hendrycks et~al.(2022)Hendrycks, Basart, Mazeika, Zou, Kwon, Mostajabi, Steinhardt, and Song}]{DBLP:conf/icml/HendrycksBMZKMS22}
Dan Hendrycks, Steven Basart, Mantas Mazeika, Andy Zou, Joseph Kwon, Mohammadreza Mostajabi, Jacob Steinhardt, and Dawn Song. 2022.
\newblock \href {https://proceedings.mlr.press/v162/hendrycks22a.html} {Scaling out-of-distribution detection for real-world settings}.
\newblock In \emph{International Conference on Machine Learning, {ICML} 2022, 17-23 July 2022, Baltimore, Maryland, {USA}}, volume 162 of \emph{Proceedings of Machine Learning Research}, pages 8759--8773. {PMLR}.

\bibitem[{Hendrycks and Gimpel(2017)}]{DBLP:conf/iclr/HendrycksG17}
Dan Hendrycks and Kevin Gimpel. 2017.
\newblock \href {https://openreview.net/forum?id=Hkg4TI9xl} {A baseline for detecting misclassified and out-of-distribution examples in neural networks}.
\newblock In \emph{5th International Conference on Learning Representations, {ICLR} 2017, Toulon, France, April 24-26, 2017, Conference Track Proceedings}. OpenReview.net.

\bibitem[{Hendrycks et~al.(2020)Hendrycks, Liu, Wallace, Dziedzic, Krishnan, and Song}]{hendrycks2020pretrained}
Dan Hendrycks, Xiaoyuan Liu, Eric Wallace, Adam Dziedzic, Rishabh Krishnan, and Dawn Song. 2020.
\newblock Pretrained transformers improve out-of-distribution robustness.
\newblock In \emph{Proceedings of the 58th Annual Meeting of the Association for Computational Linguistics}, pages 2744--2751.

\bibitem[{Hendrycks et~al.(2019)Hendrycks, Mazeika, and Dietterich}]{DBLP:conf/iclr/HendrycksMD19}
Dan Hendrycks, Mantas Mazeika, and Thomas~G. Dietterich. 2019.
\newblock \href {https://openreview.net/forum?id=HyxCxhRcY7} {Deep anomaly detection with outlier exposure}.
\newblock In \emph{7th International Conference on Learning Representations, {ICLR} 2019, New Orleans, LA, USA, May 6-9, 2019}. OpenReview.net.

\bibitem[{Hochreiter and Schmidhuber(1997)}]{hochreiter1997long}
Sepp Hochreiter and J{\"u}rgen Schmidhuber. 1997.
\newblock Long short-term memory.
\newblock \emph{Neural computation}, 9(8):1735--1780.

\bibitem[{Kamoi and Kobayashi(2020)}]{DBLP:journals/corr/abs-2003-00402}
Ryo Kamoi and Kei Kobayashi. 2020.
\newblock \href {http://arxiv.org/abs/2003.00402} {Why is the mahalanobis distance effective for anomaly detection?}
\newblock \emph{CoRR}, abs/2003.00402.

\bibitem[{Kingma and Welling(2014)}]{DBLP:journals/corr/KingmaW13}
Diederik~P. Kingma and Max Welling. 2014.
\newblock \href {http://arxiv.org/abs/1312.6114} {Auto-encoding variational bayes}.
\newblock In \emph{2nd International Conference on Learning Representations, {ICLR} 2014, Banff, AB, Canada, April 14-16, 2014, Conference Track Proceedings}.

\bibitem[{Lee et~al.(2018{\natexlab{a}})Lee, Lee, Lee, and Shin}]{DBLP:conf/iclr/LeeLLS18}
Kimin Lee, Honglak Lee, Kibok Lee, and Jinwoo Shin. 2018{\natexlab{a}}.
\newblock \href {https://openreview.net/forum?id=ryiAv2xAZ} {Training confidence-calibrated classifiers for detecting out-of-distribution samples}.
\newblock In \emph{6th International Conference on Learning Representations, {ICLR} 2018, Vancouver, BC, Canada, April 30 - May 3, 2018, Conference Track Proceedings}. OpenReview.net.

\bibitem[{Lee et~al.(2018{\natexlab{b}})Lee, Lee, Lee, and Shin}]{DBLP:conf/nips/LeeLLS18}
Kimin Lee, Kibok Lee, Honglak Lee, and Jinwoo Shin. 2018{\natexlab{b}}.
\newblock \href {https://proceedings.neurips.cc/paper/2018/hash/abdeb6f575ac5c6676b747bca8d09cc2-Abstract.html} {A simple unified framework for detecting out-of-distribution samples and adversarial attacks}.
\newblock In \emph{Advances in Neural Information Processing Systems 31: Annual Conference on Neural Information Processing Systems 2018, NeurIPS 2018, December 3-8, 2018, Montr{\'{e}}al, Canada}, pages 7167--7177.

\bibitem[{Liu et~al.(2023)Liu, Zhan, Lu, Feng, Xue, and Wu}]{liu2023good}
Bo~Liu, Liming Zhan, Zexin Lu, Yujie Feng, Lei Xue, and Xiao-Ming Wu. 2023.
\newblock How good are large language models at out-of-distribution detection?
\newblock \emph{arXiv preprint arXiv:2308.10261}.

\bibitem[{Liu et~al.(2020)Liu, Wang, Owens, and Li}]{DBLP:conf/nips/LiuWOL20}
Weitang Liu, Xiaoyun Wang, John~D. Owens, and Yixuan Li. 2020.
\newblock \href {https://proceedings.neurips.cc/paper/2020/hash/f5496252609c43eb8a3d147ab9b9c006-Abstract.html} {Energy-based out-of-distribution detection}.
\newblock In \emph{Advances in Neural Information Processing Systems 33: Annual Conference on Neural Information Processing Systems 2020, NeurIPS 2020, December 6-12, 2020, virtual}.

\bibitem[{Liu et~al.(2019)Liu, Ott, Goyal, Du, Joshi, Chen, Levy, Lewis, Zettlemoyer, and Stoyanov}]{DBLP:journals/corr/abs-1907-11692}
Yinhan Liu, Myle Ott, Naman Goyal, Jingfei Du, Mandar Joshi, Danqi Chen, Omer Levy, Mike Lewis, Luke Zettlemoyer, and Veselin Stoyanov. 2019.
\newblock \href {http://arxiv.org/abs/1907.11692} {Roberta: {A} robustly optimized {BERT} pretraining approach}.
\newblock \emph{CoRR}, abs/1907.11692.

\bibitem[{Ma et~al.(2019)Ma, Wang, Ng, Nallapati, and Xiang}]{DBLP:journals/corr/abs-1910-07973}
Xiaofei Ma, Zhiguo Wang, Patrick Ng, Ramesh Nallapati, and Bing Xiang. 2019.
\newblock \href {http://arxiv.org/abs/1910.07973} {Universal text representation from {BERT:} an empirical study}.
\newblock \emph{CoRR}, abs/1910.07973.

\bibitem[{Mohebbi et~al.(2021)Mohebbi, Modarressi, and Pilehvar}]{DBLP:conf/emnlp/MohebbiMP21}
Hosein Mohebbi, Ali Modarressi, and Mohammad~Taher Pilehvar. 2021.
\newblock \href {https://doi.org/10.18653/v1/2021.emnlp-main.61} {Exploring the role of {BERT} token representations to explain sentence probing results}.
\newblock In \emph{Proceedings of the 2021 Conference on Empirical Methods in Natural Language Processing, {EMNLP} 2021, Virtual Event / Punta Cana, Dominican Republic, 7-11 November, 2021}, pages 792--806. Association for Computational Linguistics.

\bibitem[{Morningstar et~al.(2021)Morningstar, Ham, Gallagher, Lakshminarayanan, Alemi, and Dillon}]{DBLP:conf/aistats/MorningstarHGLA21}
Warren~R. Morningstar, Cusuh Ham, Andrew~G. Gallagher, Balaji Lakshminarayanan, Alexander~A. Alemi, and Joshua~V. Dillon. 2021.
\newblock \href {http://proceedings.mlr.press/v130/morningstar21a.html} {Density of states estimation for out of distribution detection}.
\newblock In \emph{The 24th International Conference on Artificial Intelligence and Statistics, {AISTATS} 2021, April 13-15, 2021, Virtual Event}, volume 130 of \emph{Proceedings of Machine Learning Research}, pages 3232--3240. {PMLR}.

\bibitem[{Nalisnick et~al.(2019)Nalisnick, Matsukawa, Teh, G{\"{o}}r{\"{u}}r, and Lakshminarayanan}]{DBLP:conf/iclr/NalisnickMTGL19}
Eric~T. Nalisnick, Akihiro Matsukawa, Yee~Whye Teh, Dilan G{\"{o}}r{\"{u}}r, and Balaji Lakshminarayanan. 2019.
\newblock \href {https://openreview.net/forum?id=H1xwNhCcYm} {Do deep generative models know what they don't know?}
\newblock In \emph{7th International Conference on Learning Representations, {ICLR} 2019, New Orleans, LA, USA, May 6-9, 2019}. OpenReview.net.

\bibitem[{Nguyen et~al.(2015)Nguyen, Yosinski, and Clune}]{DBLP:conf/cvpr/NguyenYC15}
Anh~Mai Nguyen, Jason Yosinski, and Jeff Clune. 2015.
\newblock \href {https://doi.org/10.1109/CVPR.2015.7298640} {Deep neural networks are easily fooled: High confidence predictions for unrecognizable images}.
\newblock In \emph{{IEEE} Conference on Computer Vision and Pattern Recognition, {CVPR} 2015, Boston, MA, USA, June 7-12, 2015}, pages 427--436. {IEEE} Computer Society.

\bibitem[{Podolskiy et~al.(2021{\natexlab{a}})Podolskiy, Lipin, Bout, Artemova, and Piontkovskaya}]{DBLP:conf/aaai/PodolskiyLBAP21}
Alexander Podolskiy, Dmitry Lipin, Andrey Bout, Ekaterina Artemova, and Irina Piontkovskaya. 2021{\natexlab{a}}.
\newblock \href {https://ojs.aaai.org/index.php/AAAI/article/view/17612} {Revisiting mahalanobis distance for transformer-based out-of-domain detection}.
\newblock In \emph{Thirty-Fifth {AAAI} Conference on Artificial Intelligence, {AAAI} 2021, Thirty-Third Conference on Innovative Applications of Artificial Intelligence, {IAAI} 2021, The Eleventh Symposium on Educational Advances in Artificial Intelligence, {EAAI} 2021, Virtual Event, February 2-9, 2021}, pages 13675--13682. {AAAI} Press.

\bibitem[{Podolskiy et~al.(2021{\natexlab{b}})Podolskiy, Lipin, Bout, Artemova, and Piontkovskaya}]{podolskiy2021revisiting}
Alexander Podolskiy, Dmitry Lipin, Andrey Bout, Ekaterina Artemova, and Irina Piontkovskaya. 2021{\natexlab{b}}.
\newblock Revisiting mahalanobis distance for transformer-based out-of-domain detection.
\newblock In \emph{Proceedings of the AAAI Conference on Artificial Intelligence}, volume~35, pages 13675--13682.

\bibitem[{Ren et~al.(2019)Ren, Liu, Fertig, Snoek, Poplin, Depristo, Dillon, and Lakshminarayanan}]{ren2019likelihood}
Jie Ren, Peter~J Liu, Emily Fertig, Jasper Snoek, Ryan Poplin, Mark Depristo, Joshua Dillon, and Balaji Lakshminarayanan. 2019.
\newblock Likelihood ratios for out-of-distribution detection.
\newblock \emph{Advances in neural information processing systems}, 32.

\bibitem[{Sun et~al.(2019)Sun, Qiu, Xu, and Huang}]{DBLP:conf/cncl/SunQXH19}
Chi Sun, Xipeng Qiu, Yige Xu, and Xuanjing Huang. 2019.
\newblock \href {https://doi.org/10.1007/978-3-030-32381-3\_16} {How to fine-tune {BERT} for text classification?}
\newblock In \emph{Chinese Computational Linguistics - 18th China National Conference, {CCL} 2019, Kunming, China, October 18-20, 2019, Proceedings}, volume 11856 of \emph{Lecture Notes in Computer Science}, pages 194--206. Springer.

\bibitem[{Sun et~al.(2021)Sun, Guo, and Li}]{DBLP:conf/nips/SunGL21}
Yiyou Sun, Chuan Guo, and Yixuan Li. 2021.
\newblock \href {https://proceedings.neurips.cc/paper/2021/hash/01894d6f048493d2cacde3c579c315a3-Abstract.html} {React: Out-of-distribution detection with rectified activations}.
\newblock In \emph{Advances in Neural Information Processing Systems 34: Annual Conference on Neural Information Processing Systems 2021, NeurIPS 2021, December 6-14, 2021, virtual}, pages 144--157.

\bibitem[{Sun et~al.(2022)Sun, Ming, Zhu, and Li}]{DBLP:conf/icml/SunM0L22}
Yiyou Sun, Yifei Ming, Xiaojin Zhu, and Yixuan Li. 2022.
\newblock \href {https://proceedings.mlr.press/v162/sun22d.html} {Out-of-distribution detection with deep nearest neighbors}.
\newblock In \emph{International Conference on Machine Learning, {ICML} 2022, 17-23 July 2022, Baltimore, Maryland, {USA}}, volume 162 of \emph{Proceedings of Machine Learning Research}, pages 20827--20840. {PMLR}.

\bibitem[{Szegedy et~al.(2014)Szegedy, Zaremba, Sutskever, Bruna, Erhan, Goodfellow, and Fergus}]{DBLP:journals/corr/SzegedyZSBEGF13}
Christian Szegedy, Wojciech Zaremba, Ilya Sutskever, Joan Bruna, Dumitru Erhan, Ian~J. Goodfellow, and Rob Fergus. 2014.
\newblock \href {http://arxiv.org/abs/1312.6199} {Intriguing properties of neural networks}.
\newblock In \emph{2nd International Conference on Learning Representations, {ICLR} 2014, Banff, AB, Canada, April 14-16, 2014, Conference Track Proceedings}.

\bibitem[{Ulmer et~al.(2020)Ulmer, Meijerink, and Cin{\`a}}]{ulmer2020trust}
Dennis Ulmer, Lotta Meijerink, and Giovanni Cin{\`a}. 2020.
\newblock Trust issues: Uncertainty estimation does not enable reliable ood detection on medical tabular data.
\newblock In \emph{Machine Learning for Health}, pages 341--354. PMLR.

\bibitem[{Uppaal et~al.(2023)Uppaal, Hu, and Li}]{uppaal2023fine}
Rheeya Uppaal, Junjie Hu, and Yixuan Li. 2023.
\newblock Is fine-tuning needed? pre-trained language models are near perfect for out-of-domain detection.
\newblock In \emph{Annual Meeting of the Association for Computational Linguistics}.

\bibitem[{Vaswani et~al.(2017)Vaswani, Shazeer, Parmar, Uszkoreit, Jones, Gomez, Kaiser, and Polosukhin}]{vaswani2017attention}
Ashish Vaswani, Noam Shazeer, Niki Parmar, Jakob Uszkoreit, Llion Jones, Aidan~N Gomez, {\L}ukasz Kaiser, and Illia Polosukhin. 2017.
\newblock Attention is all you need.
\newblock \emph{Advances in neural information processing systems}, 30.

\bibitem[{Xiao et~al.(2020)Xiao, Yan, and Amit}]{xiao2020likelihood}
Zhisheng Xiao, Qing Yan, and Yali Amit. 2020.
\newblock Likelihood regret: An out-of-distribution detection score for variational auto-encoder.
\newblock \emph{Advances in neural information processing systems}, 33:20685--20696.

\bibitem[{Yang et~al.(2022)Yang, Wang, Zou, Zhou, Ding, PENG, Wang, Chen, Li, Sun, Du, Zhou, Zhang, Hendrycks, Li, and Liu}]{yang2022openood}
Jingkang Yang, Pengyun Wang, Dejian Zou, Zitang Zhou, Kunyuan Ding, WENXUAN PENG, Haoqi Wang, Guangyao Chen, Bo~Li, Yiyou Sun, Xuefeng Du, Kaiyang Zhou, Wayne Zhang, Dan Hendrycks, Yixuan Li, and Ziwei Liu. 2022.
\newblock \href {https://openreview.net/forum?id=gT6j4_tskUt} {Open{OOD}: Benchmarking generalized out-of-distribution detection}.
\newblock In \emph{Thirty-sixth Conference on Neural Information Processing Systems Datasets and Benchmarks Track}.

\bibitem[{Zhan et~al.(2021)Zhan, Liang, Liu, Fan, Wu, and Lam}]{zhan-etal-2021-scope}
Li-Ming Zhan, Haowen Liang, Bo~Liu, Lu~Fan, Xiao-Ming Wu, and Albert~Y.S. Lam. 2021.
\newblock \href {https://doi.org/10.18653/v1/2021.acl-long.273} {Out-of-scope intent detection with self-supervision and discriminative training}.
\newblock In \emph{Proceedings of the 59th Annual Meeting of the Association for Computational Linguistics and the 11th International Joint Conference on Natural Language Processing (Volume 1: Long Papers)}, pages 3521--3532, Online. Association for Computational Linguistics.

\bibitem[{Zhang et~al.(2021)Zhang, Xu, and Lin}]{zhang2021deep}
Hanlei Zhang, Hua Xu, and Ting-En Lin. 2021.
\newblock Deep open intent classification with adaptive decision boundary.
\newblock In \emph{Proceedings of the AAAI Conference on Artificial Intelligence}, volume~35, pages 14374--14382.

\bibitem[{Zhou et~al.(2021{\natexlab{a}})Zhou, Liu, and Chen}]{DBLP:conf/emnlp/Zhou0C21}
Wenxuan Zhou, Fangyu Liu, and Muhao Chen. 2021{\natexlab{a}}.
\newblock \href {https://doi.org/10.18653/v1/2021.emnlp-main.84} {Contrastive out-of-distribution detection for pretrained transformers}.
\newblock In \emph{Proceedings of the 2021 Conference on Empirical Methods in Natural Language Processing, {EMNLP} 2021, Virtual Event / Punta Cana, Dominican Republic, 7-11 November, 2021}, pages 1100--1111. Association for Computational Linguistics.

\bibitem[{Zhou et~al.(2021{\natexlab{b}})Zhou, Liu, and Chen}]{zhou2021contrastive}
Wenxuan Zhou, Fangyu Liu, and Muhao Chen. 2021{\natexlab{b}}.
\newblock Contrastive out-of-distribution detection for pretrained transformers.
\newblock In \emph{Proceedings of the 2021 Conference on Empirical Methods in Natural Language Processing}, pages 1100--1111.

\end{thebibliography}


\end{document}